%% file: main.tex
\documentclass{article} % For LaTeX2e
\usepackage{iclr2024_conference,times}

\input{math_commands.tex}

\usepackage{hyperref}
\usepackage{url}

\usepackage{graphicx}
\usepackage{booktabs}
\usepackage{multirow}
\usepackage{colortbl}
\usepackage{amsmath}
\usepackage[capitalize]{cleveref}
\usepackage{natbib}
\definecolor{Gray}{gray}{0.9}

\usepackage{xcolor}

\definecolor{Gray}{gray}{0.9}
\definecolor{LightCyan}{rgb}{0.88,1,1}

\newcolumntype{a}{>{\columncolor{Gray}}c}
\newcolumntype{b}{>{\columncolor{white}}c}

\title{mPLUG-DocOwl2: High-resolution Compressing for OCR-free Multi-page Document Understanding}

\newcommand{\pretraindataname}{MP-DocStruct1M}
\newcommand{\modelname}{DocOwl2}
\newcommand{\connectname}{H-Reducer}
\newcommand{\compresssorname}{High-resolution DocCompressor}

\iclrfinalarxiv % Uncomment this line to enable the non-anonymous mode
\begin{document}

\maketitle

\renewcommand{\thefootnote}{\fnsymbol{footnote}}
\footnotetext[1]{Corresponding author}

\begin{center}
   \centering
   \vspace{-16mm}
   \textbf{Anwen Hu\textsuperscript{1}\qquad Haiyang Xu\textsuperscript{1}\footnotemark[1] \qquad Liang Zhang\textsuperscript{2}\qquad Jiabo Ye\textsuperscript{1} \qquad Ming Yan\textsuperscript{1}\footnotemark[1]} \\ 
    \textbf{Ji Zhang\textsuperscript{1}\qquad Qin Jin\textsuperscript{2}\qquad Fei Huang\textsuperscript{1} \qquad Jingren Zhou\textsuperscript{1}} \\
     \textsuperscript{1}Alibaba Group \qquad  \textsuperscript{2}Renmin University of China \\
    {\tt\small \{huanwen.haw, shuofeng.xhy, ym119608\}@alibaba-inc.com} 
   \url{https://github.com/X-PLUG/mPLUG-DocOwl}
\end{center}

\renewcommand{\thefootnote}{\arabic{footnote}}

\begin{figure*}[h]
    \centering
    \includegraphics[width=1.0\linewidth]{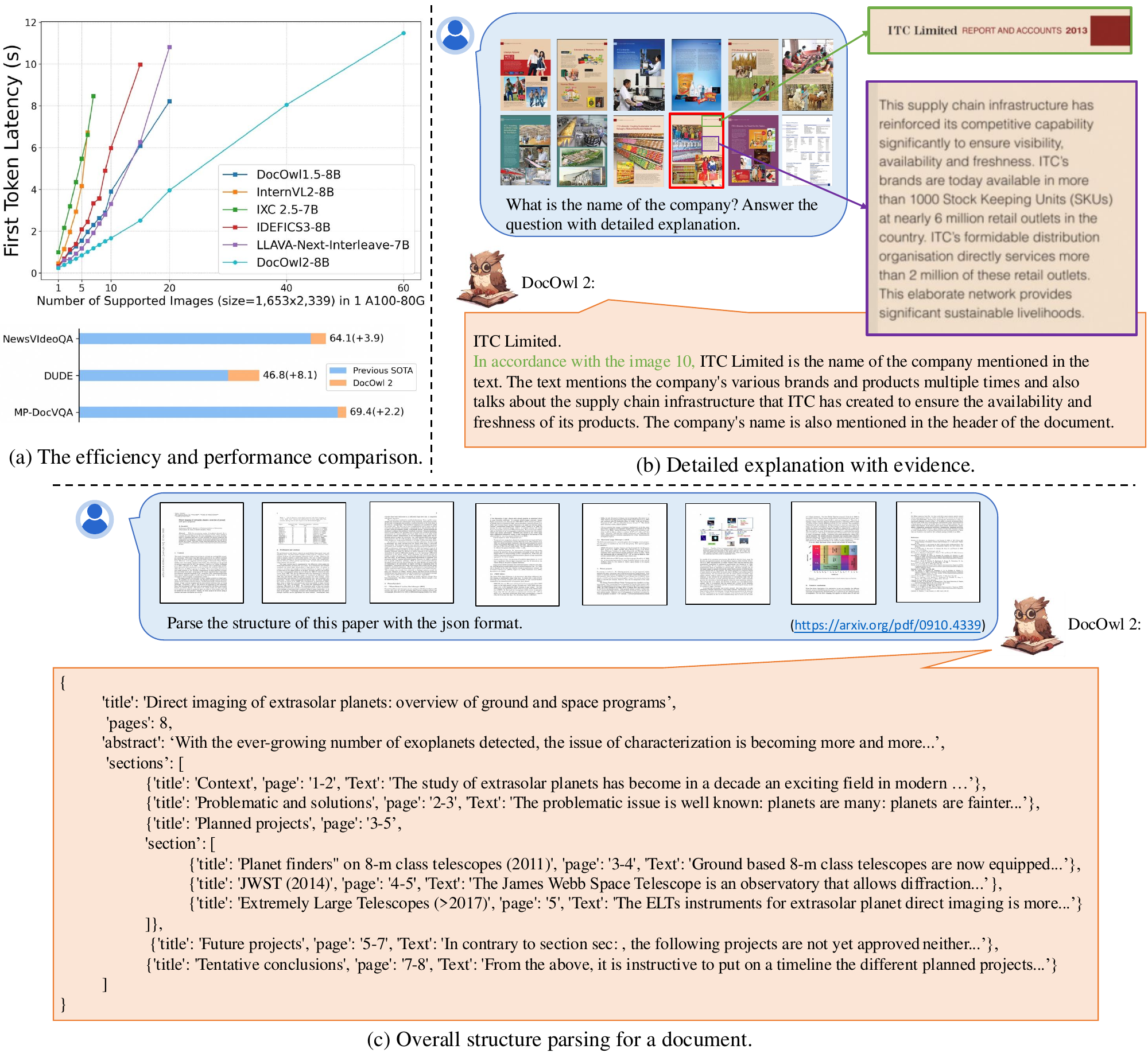}
    \caption{(a) mPLUG-\modelname~achieves state-of-the-art Multi-page Document Understanding performance with faster inference speed and less GPU memory; (b-c) mPLUG-\modelname~is able to provide a detailed explanation containing the evidence page as well as the overall structure parsing of the document.} 
    \label{fig:abs_fig}
\end{figure*}

\begin{abstract}
Multimodel Large Language Models(MLLMs) have achieved promising OCR-free Document Understanding performance by increasing the supported resolution of document images. However, this comes at the cost of generating thousands of visual tokens for a single document image, leading to excessive GPU memory and slower inference times, particularly in multi-page document comprehension. In this work, to address these challenges, 
%In this work, towards efficient OCR-free multi-page document understanding with MLLMs, 
we propose a \compresssorname~module to compress each high-resolution document image into 324 tokens, guided by low-resolution global visual features. With this compression module, to strengthen multi-page document comprehension ability and balance both token efficiency and question-answering performance, we develop the \modelname~under a three-stage training framework: Single-image Pretraining, Multi-image Continue-pretraining, and Multi-task Finetuning. \modelname~sets a new state-of-the-art across multi-page document understanding benchmarks and reduces first token latency by more than $50\%$, demonstrating advanced capabilities in multi-page questioning answering, explanation with evidence pages, and cross-page structure understanding. Additionally, compared to single-image MLLMs trained on similar data, our \modelname~achieves comparable single-page understanding performance with less than $20\%$ of the visual tokens. Our codes, models, and data are publicly available at \url{https://github.com/X-PLUG/mPLUG-DocOwl/tree/main/DocOwl2}.

\end{abstract}

\input{Introduction}

\input{RelatedWork}

\input{Method}

\input{Experiments}

\input{Conclusion}

%\newpage
\bibliography{iclr2024_conference}
\bibliographystyle{iclr2024_conference}

% \appendix
% \section{Appendix}

\end{document}

%% file: math_commands.tex
\usepackage{amsmath,amsfonts,bm}

\def\eqref#1{equation~\ref{#1}}
\def\1{\bm{1}}

\DeclareMathAlphabet{\mathsfit}{\encodingdefault}{\sfdefault}{m}{sl}
\SetMathAlphabet{\mathsfit}{bold}{\encodingdefault}{\sfdefault}{bx}{n}

%% file: Introduction.tex
\section{Introduction}
Understanding a multi-page document or news video is common in human daily life. To tackle such scenarios, Multimodal Large Language Models (MLLMs)~\citep{mplugowl,mplug-owl2,mplugowl3, qwenvl,llava} should be equipped with the ability to understand multiple images with rich visually-situated text information.
Different from natural images mainly comprising of objects, comprehending document images asks for a more fine-grained perception to recognize all texts. To tackle high-resolution document images, some works~\citep{cogagent,vary} propose to add an additional high-resolution encoder while more works~\citep{ureader, docowl1.5, internvl1.5, ixc2_4khd, ixc2.5} choose to crop a high-resolution image to low-resolution sub-images and let the Large Language Model to understand their relationship. By increasing the cropping number, the latter achieves better performance of OCR-free document understanding but also results in too many visual tokens for only 1 document image, e.g., InternVL 2~\citep{internvl1.5} costs a average of 3k visual tokens on single-page document understanding benchmark DocVQA~\citep{docvqa}. 
As shown in \cref{fig:abs_fig}(a), such long visual tokens not only result in long inference time but also occupy too much GPU memory, making it difficult to understand a complete document or video and greatly limiting their application scenarios. Inspired by Natual Language Processing work~\citep{xrag, icae,auto_compressor} which summarizes a textual paragraph/document into fewer tokens and maintains most semantics, we argue that visual tokens of document images can also be further compressed while maintaining both layout and most textual information. 

\begin{figure*}[tp]
    \centering
    \includegraphics[width=1.0\linewidth]{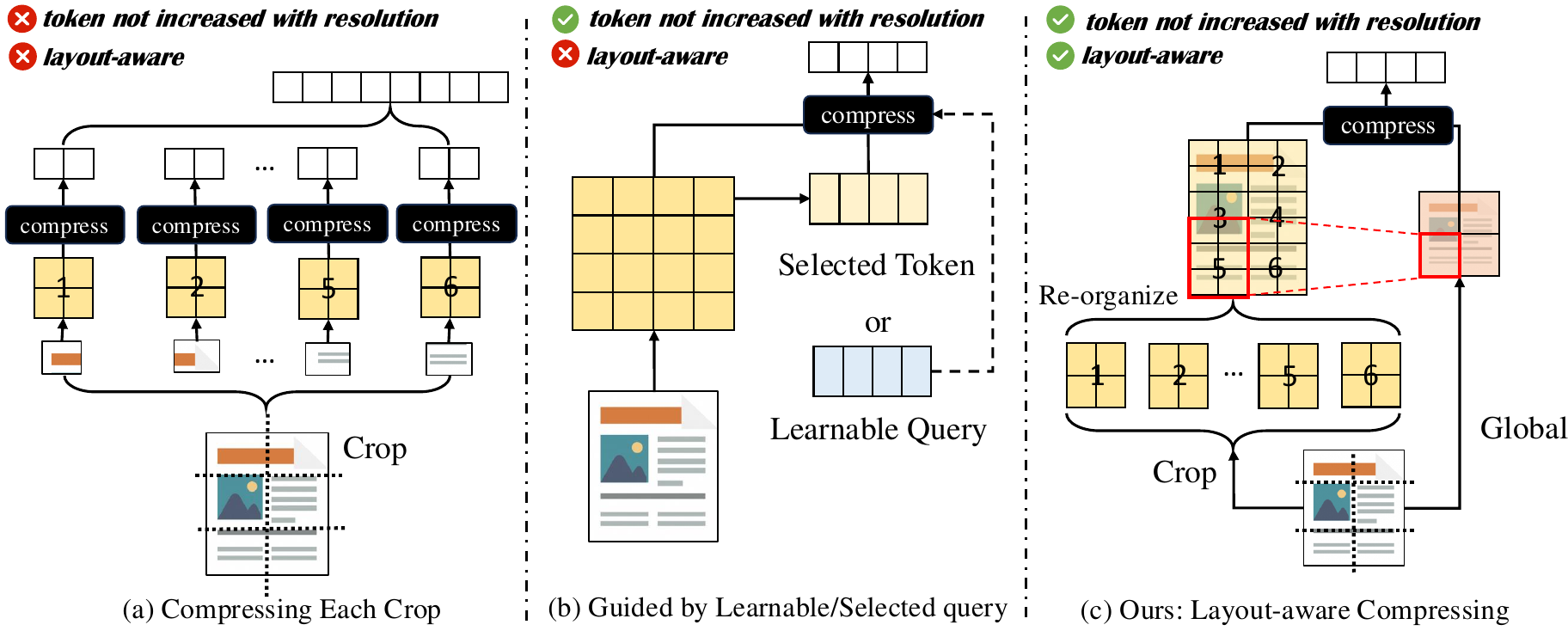}
    \caption{Illustrations of different compressing methods for OCR-free document understanding.} 
    \label{fig:intro}
\end{figure*}

Existing compressing architecture in MLLMs are hard to balance information retention and token efficiency during document image encoding. As shown in \cref{fig:intro}(a), independently compressing each crop of a document image~\citep{tokenpacker, docowl1.5} could reduce visual tokens of each sub-image but still results in a long sequence of visual tokens after concatenating all sub-images. Leveraging learnable queries~\citep{qwenvl,blip2,mplugowl} or selected tokens~\citep{textmonkey} as compressing guidance could produce an identical length of tokens for any resolution but overlook the overall layout information, as shown in \cref{fig:intro}(b). Layout-aware guidance is important for compressing visual features of document images because texts within a layout region are semantic-coherent and easier to summarize. For example, in a two-column paper, texts belonging to the `Related Work' section are difficult to summarize with texts on the same line but belonging to the `Method' section.

 In this work, as shown in \cref{fig:intro}(c), we propose a layout-aware compressing architecture \textbf{\compresssorname}~based on cross-attention to compress document images into fewer tokens and achieve better performance than existing compressing methods. Considering that a global low-resolution image can well capture the overall layout information, we utilize visual features of a global low-resolution image as the compressing guidance (query). Each visual feature in the global feature map just captures the layout information of partial regions. Therefore, each query attending to all high-resolution features will not only make information compression more difficult but also increase computation complexity. To summarize text information within a layout region, for each query from the global feature map, a group of high-resolution features with identical relative positions in the raw image is collected as compressing objects, sometimes spanning multiple sub-images. Besides, since the vision-to-text (V2T) module of MLLMs could convert visual features into textual feature space, we argue that compressing visual features after the vision-to-text module could better maintain textual semantics in document images. Therefore, based on the architecture of DocOwl 1.5~\citep{docowl1.5}, we propose mPLUG-\modelname~by placing the \compresssorname~afther its V2T module: H-Reducer. To take full advantage of the compressing method, our model \modelname~is trained with a three-stage framework: Single-image Pretraining, Multi-image Continue-Pretraining, and Multi-task Finetuning to support both single-image and multi-image/frame understanding. Our experiments on single-page and multi-page document benchmarks demonstrate the good balance of OCR-free document understanding performance and token efficiency of \modelname. We perform sufficient ablation studies to validate the superiority of our \compresssorname~and the benefits of the three-stage training framework for both single-page and multi-page understanding performance.

 Our contributions in this work are three-fold:
 \begin{itemize}
    \item We propose a novel compressing architecture, namely \compresssorname, to greatly reduce visual tokens of high-resolution document images. Compared with existing compressing methods, our method achieves better OCR-free single-image document Understanding performance with fewer visual tokens.
     \item \modelname~achieves state-of-the-art performance on Multi-page Document understanding benchmarks with $\textless 50\%$ First Token Latency.
    \item Compared with state-of-the-art MLLMs with similar model size and training data, \modelname~ achieves comparable performance with $\textless 20\%$ visual tokens on 10 single-image document benchmarks.
   
\end{itemize}

%% file: RelatedWork.tex
\section{Related Work}

\subsection{OCR-free Visual Document Understanding}
Visual Document Understanding aims to comprehend images with rich text information, including scans of document pages~\citep{docvqa,mpdocvqa,dude,mpmqa, vary}, infographics~\citep{infovqa}, charts~\citep{chartqa,dvqa,plotqa,figureqa}, tables images~\citep{wikitableqa,TabFact,pubtabnet}, webpage screenshots~\citep{visualmrc,websrc} and natural images with scene texts~\citep{textvqa,textcaps, qctextcap}. Recently, many Multimodal Large Language Models have been proposed to perform visual document understanding in an OCR-free manner. mPLUG-DocOwl~\citep{docowl} and UReader~\citep{ureader} first propose to unify different tasks across 5 types of document images in the seq-to-seq format. To encode rich text information in high-resolution images, UReader~\citep{ureader} proposes a Shape-adaptive Cropping Module to cut the raw image into multiple low-resolution sub-images and utilizes an identical low-resolution encoder to encode both sub-images and a global image. Monkey~\citep{monkey} proposes to employ a sliding window to partition high-resolution images and a resampler to reduce redundant information of each sub-image. mPLUG-DocOwl1.5~\citep{docowl1.5} increases the basic resolution of the low-resolution encoder and replaces the Visual Abstractor~\citep{mplugowl} with 1 simple convolution layer to better maintain the structure information. DocPedia~\citep{docpedia} directly processes high-resolution images in the frequency domain. CogAgent~\citep{cogagent} proposes to utilize a high-resolution encoder to encode high-resolution visual features and a low-resolution encoder to encode low-resolution global features. Series work of InternLM-XComposer~\citep{ixc2.5, ixc2_4khd} and InternVL~\citep{internvl1.5} further optimize the cropping method or increase the cropping number and greatly improves the OCR-free Document Understanding performance. These works achieve promising performance but suffer from too many visual tokens for a high-resolution image (always $\textgreater 1k$ tokens for a common A4-sized document page), which hinders the development of OCR-free multi-page document understanding.

\subsection{Visual Feature Compressing}
Reducing visual tokens of a single image enables a Multimodal Large Language Model with limited maximum sequence length to leverage more images as contexts to perform complex multimodal tasks, such as video understanding, embodied interaction, or multi-page document understanding. There have been some architectures proposed for compressing visual features of general images with fewer learnable queries, such as the Resampler~\citep{Alayrac2022FlamingoAV,qwenvl}, Abstractor~\citep{mplugowl,mplug-owl2} and Q-former~\citep{blip2}. Randomly initialized Learnable queries can ensemble object information in general images but is hard to summarize rich text information in high-resolution document images. As a compromise solution, TokenPacker~\citep{tokenpacker} proposes to compress each sub-image with its downsampled visual features as the query to perform cross-attention. TokenPacker just reduces each sub-image's visual tokens, thus still creates more than 1k visual tokens when processing high-resolution document images. TextMonkey~\citep{textmonkey} first filters valuable visual tokens and then uses them as guidance to aggregate all visual tokens. Due to that valuable visual tokens are selected by measuring the token similarity, visual information of partial regions may not be covered and thus not well compressed during following cross-attention. In this work, our \compresssorname~leverages visual features from the row-resolution global images as the query, the ensembled feature map of sub-images as key and value. This not only produces a fixed number of visual tokens for images of any resolution but also covers all areas during compression. Compared to Mini-Gemini~\citep{mini-gemini} which compresses general visual features, there are major two differences with our \modelname. Firstly, we make full use of global visual features and sub-image features produced by an identical low-resolution vision encoder and don't need to add an extra high-resolution encoder. Secondly, for better summarizing textual information in document images, our cross-attention is applied based on visual features that have been aligned with textual features of LLM. We argue that directly compressing outputs of the vision encoder will lose more visually situated textual information while comprising features aligned with LLM is like summarizing texts~\citep{xrag, icae,auto_compressor} and can better maintain textual semantics in document images. Fair comparisons are performed in our experiments to support our hypothesis.

%% file: Method.tex
\section{mPLUG-\modelname}

\begin{figure*}[tp]
    \centering
    \includegraphics[width=0.9\linewidth]{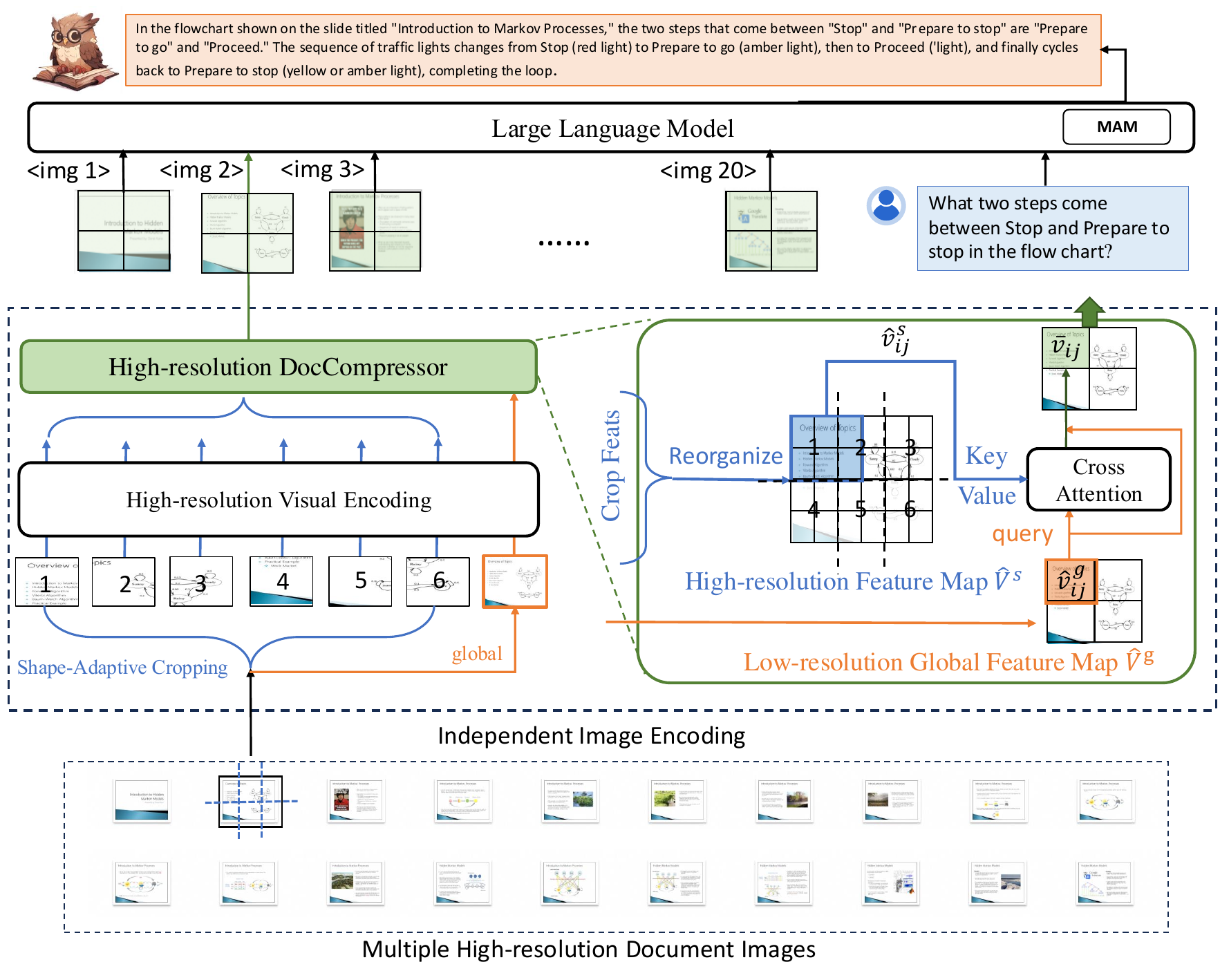}
    \caption{The architecture of \modelname. Each image is independently encoded by the pipeline of Shape-adaptive Cropping, High-resolution Visual Encoding and \compresssorname.} 
    \label{fig:model}
\end{figure*}

As shown in \cref{fig:model}, \modelname~leverages a Shape-adaptive Cropping Module and a low-resolution vision encoder to encode high-resolution document images. Then, it utilizes a vision-to-text module \connectname~to ensemble horizontal visual features and align the dimension of vision features with Large Language Models. Furthermore, a high-resolution compressor is designed to greatly reduce the number of visual features while maintain most visual information. Finally, compressed visual tokens of multiple images/pages are concatenated with text instructions and input to a Large Language Model for multimodal understanding.

\subsection{High Resolution Vision Encoding}
Following UReader~\citep{ureader} and DocOwl 1.5~\citep{docowl1.5}, \modelname~utilizes a parameter-free Shape-adaptive Cropping Module to preprocess high-resolution images. Concretely, it cuts each high-resolution image $I$ into $R \times C$ size-fixed sub-images $I^{s} = \{I^{s}_{xy}\}, 1 \leq x \leq R, 1 \leq y \leq C$, where cropping rows $R$ and columns $C$ are flexibly decided based on the raw resolution of $I$. Besides, to maintain the overall layout information, the raw image is also directly resized to a global image $I^g$. Both the global image and sub-images are sized $H \times W$. 

After the cropping module, a low-resolution transformer-based vision encoder ViT~\citep{vit2021} is utilized to independently extract vision features of each sub-image and the global image as follows:

 \begin{gather}
    V^{g} = {\rm ViT}(I^g) \\
    V^{s}_{xy} = {\rm ViT}(I^{s}_{xy}), 1 \leq x \leq R, 1 \leq y \leq C,
\end{gather}
where both $V^{g}$ and $V^{s}_{xy}$ are visual features with the shape of $h \times w \times d$, $d$ is the feature dimension and $w, h$ are the width and height of the feature map.

Following DocOwl 1.5, after the ViT, for each sub-image or global image, we apply a vision-to-text module \connectname~ to ensemble horizontal 4 features by a convolution layer and align the feature dimension with the Large Language Model with a fully connected layer. The calculation of \connectname~is represented as follows:
 \begin{gather}
    \hat{V} = {\rm FC}({\rm Conv}(V)), V \in \{V^{g}, V^{s}_{xy}\}, 1 \leq x \leq R, 1 \leq y \leq C,
\end{gather}
where the shape of the visual feature map $\hat{V}$ is $h \times \frac{w}{4} \times \hat{d}$, $\hat{d}$ is the dimension of hidden states of the large language model.

\subsection{High Resolution Full-Compressing}
Although the \connectname~has reduced the visual tokens of each sub-image or global image to $\frac{1}{4}$ the length of original visual features, the token length of high-resolution images is still too long to perform multi-page/image joint understanding for Large Language Models. For example, the token 
 length of 1 high-resolution image in DocOwl 1.5~\citep{docowl1.5} is $(R \times C+1) \times h \times \frac{w}{4}$, which will be 2,560 when the raw resolution is $1,344 \times 1,344$. 

In Natural Language Processing, a sentence/paragraph/document of text tokens can be compressed into fewer summary vectors while maintaining most semantics~\citep{xrag, icae,auto_compressor}. Besides, since visual features have been aligned with the textual feature space of large language models, the visual tokens of document images after the vision-to-text module can also be treated as textual tokens encoding different parts of textual information in the image. Thus, taking into account these two points, in this work, we argue that visually situated textual information of document images can also be further compressed into fewer tokens, especially after the vision-to-text alignment.

Ideally, the compression of visual texts should be based on their layout. Texts from the same layout region (e.g., a title/paragraph region) are more appropriate to be fused into an identical token. After the vision-to-text module \connectname, the global visual feature $\hat{V}^{g}$ mainly encodes the overall text layout information while visual features of sub-images $\{\hat{V}^{s}_{xy}\}$ capture detailed textual information. Besides, due to both the global image and cropped sub-images come from an identical image, there is a clear mapping between the visual tokens of $\hat{V}^{g}$ and $\{\hat{V}^{s}_{xy}\}$. As shown in \cref{fig:model}, each visual token in $\hat{V}^{g}$ can be aligned with $ R \times C $ visual tokens in $\{\hat{V}^{s}_{xy}\}$. Therefore, in this work, with global visual features as query, and the visual features from sub-images as key and value, we propose to utilize cross-attention to ensemble textual semantics and greatly reduce the number of visual tokens of a high-resolution image to the one of a low-resolution global image. 

Concretely, we first re-organize feature maps of cropping images ($\{\hat{V}^{s}_{xy}\}, 1 \leq x \leq R, 1 \leq y \leq C$) to a complete feature map $\hat{V}^{s}$ according to their positions in the raw high-resolution image. Then, for each visual token in the feature map $\hat{V}^{g}$ of the global image, we collect its corresponding $R \times C$ visual tokens from $\hat{V}^{s}$ as the key and value, the cross-attention layer in this compressor is calculated as follows:

\begin{gather}
    \hat{v}^g_{ij} \in \hat{V}^g, 1 \leq i \leq h, 1 \leq j \leq w/4\\ 
    \hat{v}^s_{ij} = [\hat{v}^s_{i^{'}j^{'}}] \subset \hat{V}^s, (i-1)R+1 \leq i^{'} \leq iR, (j-1)C+1 \leq j^{'} \leq jC  \label{equ:group_att} \\ 
    \bar{v}_{ij} = {\rm softmax}(\frac{W^{q}\hat{v}^g_{ij}W^{k}\hat{v}^s_{ij}}{\sqrt{d_k}})W^{v}\hat{v}^s_{ij} + \hat{v}^g_{ij}
\end{gather}

where $\hat{v}^g_{ij}$ is a visual token from the feature map of the global image, $\hat{v}^s_{ij}$ are visual tokens from the re-organized feature map of cropping images. $\hat{v}^g_{ij}$ and $\hat{v}^s_{ij}$ correspond to the same area in the raw image. $W^q,W^k,W^v$ are learnable projection matrics. 

After high-resolution compressing, the compressed feature map of each image is organized into a sequence $\bar{V}=[\bar{v}_1,\bar{v}_2,...,\bar{v}_{h \times \frac{w}{4}}]$  for subsequent understanding of the large language model.

\subsection{Multi-image Modeling with LLM}
Through the high-resolution compressing, the number of visual tokens for each high-resolution image is reduced from  $(R \times C+1) \times h \times \frac{w}{4}$ to $h \times \frac{w}{4}$. Such efficient vision encoding allows joint understanding of multiple document images with Large Language Models. To help the LLM better distinguish visual features from different images and understand the ordinal number of images, we add a textual ordinal token \texttt{`<img $x$>'} before the visual features of each image, where $x$ is the ordinal number. Overall, the decoding of the decoder for multiple images is as follows:
\begin{gather}
    Y = {\rm LLM}([P_0;\bar{V}_0; P_1;\bar{V}_1, ...,P_n; \bar{V}_n;T])
\end{gather}
where $[;]$ means the concatenation operation, $n$ is the number of images, $P_x, 1 \leq x \leq n$ is the textual embedding of the ordinal token \texttt{`<img $x$>'},  $\bar{V}_x$ is the visual features for each image, $T$ is the textual instruction and $Y$ is the predicted answer. 

\subsection{Model Training}
\modelname~is trained with three stages: Single-image Pre-training, Multi-image Continue Pretraining, and Multi-task Finetuning.

At the first stage, to ensure the compressed visual tokens can encode most visual information, especially visually situated texts, we first perform Unifed Structure Learning as DocOwl 1.5 with the dataset DocStruct4M~\citep{docowl1.5}, which covers the learning of struct-aware document parsing, table parsing, chart parsing and natural image parsing of a single image. 

After Single-image Pretraining, to empower our model with the ability to correlate multiple images, we further perform Multi-image Continue Pretraing with a struct-aware multi-page document parsing dataset \pretraindataname. With partial documents from two datasets of PixParse\footnote{\url{https://huggingface.co/datasets/pixparse/idl-wds}}\footnote{\url{https://huggingface.co/datasets/pixparse/pdfa-eng-wds}}, we design two symmetrical tasks of multi-image understanding: Multi-page Text Parsing and Multi-page Text Lookup. Given successive page images in a document, the Multi-page Text Parsing instructs the model to parse texts of specified one or two pages, such as \texttt{`Recognize texts in image 2 and image 10.'}. As for the Multi-page Text Lookup task, with texts from 1-2 pages as input, the model is required to predict the concrete ordinal number of images containing these texts, for example, \texttt{`Looking for the image with text <doc> ...</doc> and <doc> ...</doc>.'}. 
Besides \pretraindataname, during this stage, we also randomly chose 0.5M samples from DocStruct4M to avoid the catastrophic forgetting of structure parsing across different types of images.

Finally, we ensemble single-image and multi-image instruction tuning datasets to perform multi-task tuning. We leverage DocDownstream-1.0~\citep{docowl1.5} and DocReason25K~\citep{docowl1.5} as single-image datasets. DocDownstream-1.0 is an ensembled dataset comprising of DocVQA~\citep{docvqa}, InfoVQA~\citep{infovqa}, DeepForm~\citep{deepform}, KLC~\citep{klc}, WTQ~\citep{wikitableqa}, TabFact~\citep{TabFact}, ChartQA~\citep{chartqa}, TextVQA~\citep{textvqa}, TextCaps~\citep{textcaps} and VisualMRC~\citep{visualmrc}. DocReason25K is a question-answering dataset with detailed explanations.
As for multi-image understanding, we ensemble 2 document datasets, MP-DocVQA~\citep{mpdocvqa} and DUDE~\citep{dude}, and 1 news video dataset NewsVideoQA~\citep{newsvideoqa} as concise question-answering datasets. MP-DocVQA contains 46k question-answering pairs on 60k page images scanned from 6k industry documents with rich tables, diagrams, pictures, and both handwritten and printed texts. DUDE covers more domains of documents, including medical, legal, technical, financial, etc. It contains 41k question-answering pairs on 5k documents. NewsVideoQA collects news videos with rich visually-situated texts from diverse English news channels around the world, such as BBC, CNN, etc. It contains 8k question-answering pairs framed on 3k videos. Besides, to trigger the ability of detailed explanations with evidence pages, we built MP-DocReason51K based on DocReason25K. Concretely, for each single-image sample from DocReason25K, we construct two multi-image samples with noisy images randomly chosen from the same or different categories. After randomly inserting the evidence image into noisy images, we add an extra evidence description (e.g., \texttt{`According to the 5th image,'}) into the raw detailed explanation to get the target of multi-image samples. Most question-answering samples just focus on 1-2 pages of a document, to further strengthen the ability of a comprehensive understanding of a document, we leverage a small part of annotations from DocGenome~\citep{docgenome} to construct text sequences in the JSON format, which represents the hierarchical structure of a scientific paper and partial detailed texts.

The detailed statistics of training datasets of \modelname~are shown in \cref{tab:dataset_statistic}.

\begin{table*}
    \caption{Detailed statistic of training datasets of \modelname.}
    \label{tab:dataset_statistic}
    \footnotesize
    % \small
    \centering
    \begin{tabular}{cccc}
    \toprule
    \textbf{Training Stage} & \textbf{Input Image} & \textbf{Dataset} & \textbf{Num} \\
    \toprule
    Single-image Pretraining & Single & DocStruct4M & 4,036,402 \\
    \midrule
    \multirow{2}*{Multi-image Continue Pretraining} & Single & DocStruct4M & 501,781 \\
    ~ & Multiple & MP-DocStruct1M & 1,113,259 \\
    \midrule
    \multirow{5}*{Multi-task Finetuning} & \multirow{2}*{Single} & \begin{tabular}[c]{@{}l@{}}DocVQA, InfoVQA, DeepForm, \\  KLC, WTQ, TabFact, ChartQA, \\ TextVQA, TextCaps, VisualMRC \end{tabular} & 552,315 \\
    ~ & ~ & DocReason25K  & 25,877 \\
    \cline{2-4}
    ~ & \multirow{5}*{Multiple} & MP-DocVQA  & 70,154  \\
    ~ & ~ & DUDE  & 35,438  \\
    ~ & ~ & NewsVideoQA  & 8,619 \\
    ~ & ~ & MP-DocReason51K & 51,754 \\
    ~ & ~ & DocGenome12K & 12,010 \\
    \midrule
    \end{tabular}
    
\end{table*}

%% file: Experiments.tex
\section{Experiments}
\vspace{-8pt}
\subsection{Implementation Details}
The maximum number of crops is set to 12. The resolution of each sub-image or the global image is 504x504. The \compresssorname~comprises of 2 layers of cross attention. Initialized from mPLUG-Owl2~\citep{mplug-owl2}, the vision encoder (ViT/L-14~\citep{vit2021}), H-Reducer and \compresssorname~are trained during the Sinlge-image Pretraining. Besides, the main parameters of the Large Language Model~\citep{llama} are frozen while a Modality Adaptive Module (MAM)~\citep{mplug-owl2} used to distinguish visual and textual features in the LLM is tuned. The first stage is trained 12k steps with a batch size of 1,024 and the learning rate set as 1e-4. During the Multi-image Continue-pretraining, the vision encoder is further frozen and the H-Reducer, \compresssorname~and MAM is tuned. The second stage is trained 2.4k steps with a batch size of 1,024 and the learning rate set as 2e-5. At the final Multi-task Finetuning stage, all parameters except the vision encoder are optimized. The batch size, training step, and learning rate at this stage are set as 256, 9k, and 2e-5, respectively.

\subsection{Main Results}

\begin{table*}
    \caption{Comparison with OCR-free methods on single-image document understanding tasks. The `$*$' refers to models without LLMs and separately fine-tuned on each downstream task. `Token$^V$' means the average number of visual tokens of a single image. `\textbf{Bold}' means SOTA performance within the group and `\underline{Underline}' means achieving 80\% SOTA performance among all baselines.}
    \label{tab:main}
    \footnotesize
    %\small
    \centering
    \resizebox{\linewidth}{!}{
    \begin{tabular}{c|lcc|cccc|cc|c|cc|c}
    \toprule
     ~ & \multirow{2}*{\textbf{Model}} & \multirow{2}*{\textbf{Size}} & \multirow{2}*{\textbf{Token$^V$}} & \textbf{Doc} & \textbf{Info} & \textbf{Deep} & \multirow{2}*{\textbf{KLC}} & \multirow{2}*{\textbf{WTQ}}  & \textbf{Tab} & \textbf{Chart} & \textbf{Text} & \textbf{Text} & \textbf{Visual} \\ 
    ~ & ~ & ~  & ~  & \textbf{VQA} & \textbf{VQA} & \textbf{Form} & ~ & ~ & \textbf{Fact} & \textbf{QA} & \textbf{VQA} & \textbf{Caps} & \textbf{MRC} \\
    \midrule
     ~ & Donut$^{*}$ & \textless 1B & 4,800 & 67.5 & 11.6 & 61.6 & 30.0 & 18.8 & 54.6 &41.8 & 43.5 & 74.4 & 93.91 \\
    ~ & Pix2Struct$_{base}^{*}$ & \textless 1B & 2,048 &  72.1 & 38.2 &- & - & - & - & 56.0 & -& 88.0 & -  \\ 
    ~ & Pix2Struct$_{large}^{*}$ & 1B & 2,048 & 76.6 & 40.0 & - & - & - & - & 58.6 & -& 95.5 & -  \\ 
    \midrule
    \multirow{7}*{\rotatebox{90}{$Token^V \geq 1k$}} & CogAgent & 17B & 6,656 & 81.6 & 44.5 & -& -&- & - & 68.4 & 76.1 & - & -  \\
    ~ & IXC 2.5 & 7B & $\sim$ 5,118 & 90.9 & 69.9 & \textbf{71.2} & -& \textbf{53.6} & \textbf{85.2} & 82.2 & 78.2 & - & \textbf{307.5} \\
    ~ &InternVL 2 & 8B & $\sim$ 3,133 & \textbf{91.6}  & \textbf{74.8} & - & - & - & - & \textbf{83.3} & \textbf{77.4} &  \\
    ~ &TokenPacker & 13B & $\sim$ 1,833 & 70.0 & - & - & - & - & - & - & - & - & -  \\
    ~ &DocOwl 1.5 &8B  & $\sim$ 1,698  & 82.2 & 50.7 & 68.8 & \textbf{38.7} & 40.6  & 80.2 & 70.2 & 68.6 & \textbf{131.6} & 246.4 \\
    ~ &DocPeida & 7B & 1,600 & 47.1 & 15.2 & - & - & - & - & 46.9 & 60.2 & - & - \\
    ~ &Monkey & 9B & 1,280 & 66.5 & 36.1 & 40.6 & 32.8 & 25.3 & - & - & 64.3 & 93.2 & - \\
    \midrule
    \multirow{6}*{\rotatebox{90}{$Token^V < 1k$}} & DocOwl & 7B & $\sim$ 841 & 62.2 & 38.2 & 42.6 & 30.3 & 26.9 & 60.2 & 57.4 & 52.6 & 111.9 & 188.8 \\
    ~ &UReader &7B & $\sim$841 & 65.4 & 42.2 & 49.5 & 32.8 & 29.4  & 67.6 & 59.3 & 57.6 & 118.4 & \textbf{221.7} \\
    ~ &TextMonkey & 9B & 768 & 73.0 & 28.6 & 59.7 & \textbf{37.8} & 31.9 & - & 66.9 & 65.9 & - & - \\
    ~ &TokenPacker & 13B & $\sim$ 467 & 58.0  & - & - & - & - & - & - & - & - & -\\
    ~ &QwenVL &9B & 256 & 65.1 & 35.4 & - & - & - &-& 65.7 & 63.8 & - & -  \\
    ~ &Vary &7B & 256 & 76.3 & - & - & - & - & - & 66.1 & - & - & -  \\
    ~ &\modelname & 8B & 324 & \textbf{\underline{80.7}} & \textbf{46.4} & \textbf{\underline{66.8}} & \underline{37.5} & \textbf{36.5}  & \textbf{\underline{78.2}} & \textbf{\underline{70.0}} & \textbf{\underline{66.7}} & \textbf{\underline{131.8}} & 217.4 \\
    \bottomrule
    \end{tabular}
    }
\end{table*}

We compare \modelname~with state-of-the-art Multimodal Large Language Models on 10 single-image document understanding benchmarks, 2 Multi-page document Understanding benchmarks, and 1 text-rich video understanding benchmark. Both question-answering performance and the First Token Latency (seconds) are considered to show the effectiveness of our model. 

\subsubsection{Single-image Document Understanding}
\begin{figure}[tp]
\begin{center}
    \includegraphics[width=0.9\linewidth]{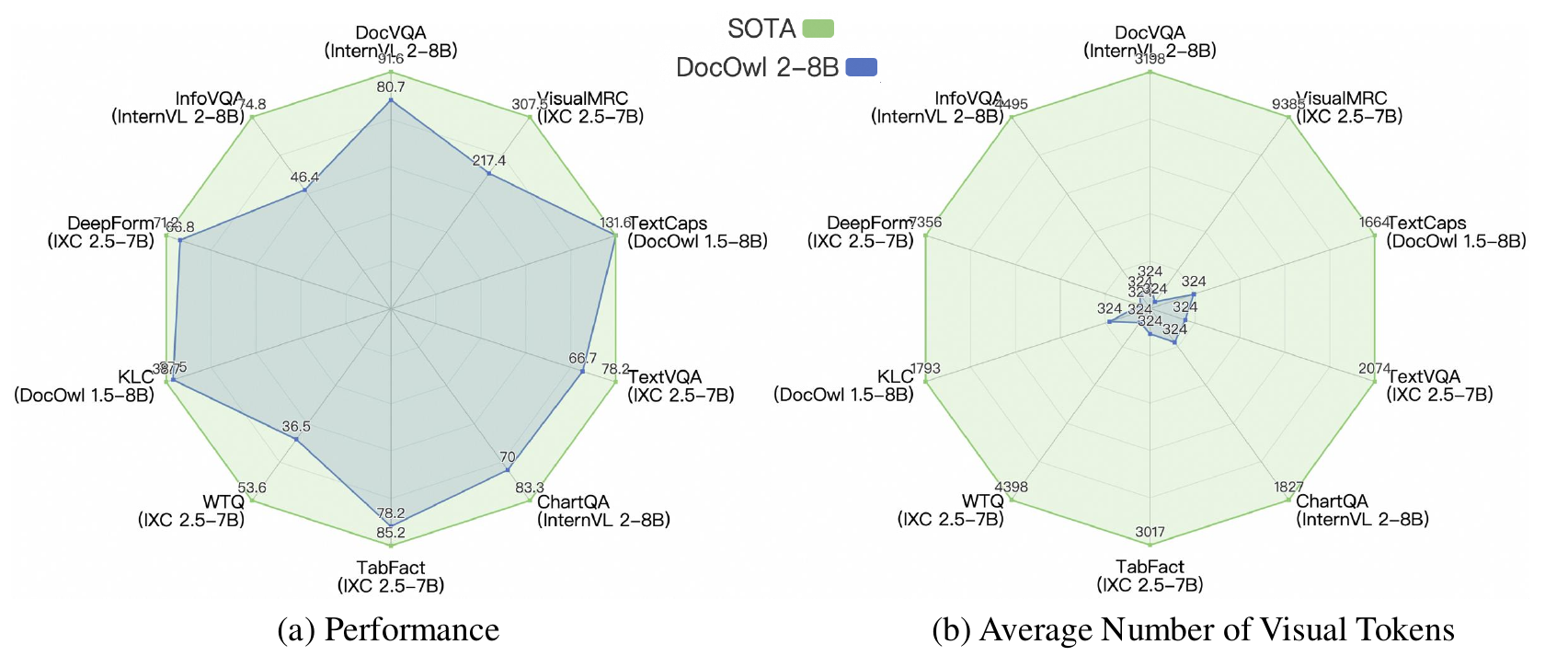}
\end{center}
\vspace{-8pt}
\caption{The comparison of our \modelname~with state-of-the-art Multimodal Large Language Models on (a) OCR-free performance and (b) the average number of visual tokens on 10 Visual Document Understanding benchmarks.}
\label{fig:radar}
\end{figure}

For Single-image Document Understanding, we divide baselines into three groups: (a) models without Large Language Models as decoders~\citep{donut, pix2struct}, (b) Multimodal LLMs~\citep{cogagent,ixc2.5, internvl1.5, tokenpacker, docowl1.5, docpedia, monkey} with an average number of visual tokens over 1k for a single document image and (c) Multimodal LLMs~\citep{docowl, ureader, textmonkey, tokenpacker, qwenvl} with an average number of visual tokens less than 1k. As shown in \cref{tab:main}, although specifically fine-tuned on each downstream dataset, Donut~\citep{donut} or PixsStruct~\cite{pix2struct} are not as good as Multimodal LLMs, showing the potential of MLLMs for generalized OCR-free document understanding. Compared with MLLMs with $\textless 1k$ visual tokens, our \modelname~achieves better or comparable performance on 10 benchmarks. Especially, with fewer visual tokens, our model outperforms both TextMonkey~\citep{textmonkey} and TokenPacker~\citep{tokenpacker} which also aim to compress visual tokens, showing that our layout-aware architecture \compresssorname~is better at summarizing and maintaining textual information in high-resolution document images. Besides, compared with state-of-the-art MLLMs with $\textgreater 1k$ visual tokens, \modelname~achieves $\textgreater 80\%$ performance on 7/10 benchmarks while with $\textless 20\%$ visual tokens. \cref{fig:radar} visualizes the comparison with SOTA in terms of question-answering performance and the number of visual tokens. 

\begin{table*}[tp]
    \caption{Comparison with OCR-free Multimodal Large Language Models on single-image document understanding benchmarks. `FTL(s)' refers to the First Token Latency (seconds)}
    \label{tab:single_ftl}
    \footnotesize
    %\small
    \centering
    \resizebox{\linewidth}{!}{
    \begin{tabular}{lc|cac|cac|cac}
    \toprule
    \multirow{2}*{\textbf{Model}} & \multirow{2}*{\textbf{Size}}  & \multicolumn{3}{c|}{\textbf{DocVQA}} & \multicolumn{3}{c|}{\textbf{ChartQA}} & \multicolumn{3}{c}{\textbf{TextVQA}} \\ 
    ~ & ~  & Token$^V$ & FTL(s)$\downarrow$ & ANLS$\uparrow$ & Token$^V$ & FTL(s)$\downarrow$ & ANLS$\uparrow$ & Token$^V$ & FTL(s)$\downarrow$ & ANLS$\uparrow$ \\
    \toprule
    InternVL 2 & 8B & $\sim$ 3,198 & 0.94 & 91.6 & $\sim$ 1,827 &  0.56 & 83.3 & $\sim$2,864 & 1.01 & 77.4 \\
    IXC 2.5 & 7B & $\sim$7,395&   3.73 & 90.9 & $\sim$1,971 &  1.05 & 82.2 & $\sim$2,075 &   1.11 & 78.2 \\
    DocOwl 1.5 & 8B & $\sim$1,806 &  0.58 & 82.2 & $\sim$1,713 &  0.53 & 70.2 & $\sim$1,664 &  0.56 & 68.6 \\
    \midrule
    TextMonkey & 9B & 768 & 0.58 & 73.0 & 768 &  0.51 & 66.9  & 768 &  0.50 & 65.9 \\
    \modelname & 8B & 324 &  \textbf{0.26} & 80.7 & 324 &  \textbf{0.21} & 70.0  & 324 &  \textbf{0.23} & 66.7 \\
    \midrule
    \end{tabular}
    }
\end{table*}

Furthermore, we compare the First Token Latency (seconds) on the 3 most frequently compared datasets, representing documents, charts, and natural images. As shown in \cref{tab:single_ftl}, the far greater number of visual tokens enable InternVL 2~\citep{internvl1.5} and IXC 2.5~\citep{ixc2.5} to achieve better performance but also result in higher inference time. Considering the model architecture and training data, it's most fair to compare \modelname~with DocOwl 1.5. After adding the \compresssorname, with similar training data of OCR learning, \modelname~achieves 98\% performance of DocOwl 1.5 while reducing 50\% First Token Latency with just 20\% visual tokens. This validates the effectiveness of our compressor for compressing visually-situated text information on the most common documents, charts, and natural images.

\subsubsection{Multi-page/Video Document Understanding}
For Multi-page Document Understanding and Text-rich Video Understanding benchmarks, we choose recently proposed Multimodal LLMs~\citep{longva,idefics,llava-next-interleave} with multi-page OCR-free document understanding abilities and can be fed into more than 10 images under a single A100-80G as baselines. As shown in \cref{tab:multi-image}, with fewer visual tokens for a single image/frame, our model \modelname~achieve better question-answering performance and much less First Token Latency, validating the good balance of \modelname~between the OCR-free document understanding performance and token efficiency.

\begin{table*}
    \caption{Comparison with OCR-free Multimodal Large Language Models on multi-image/video document understanding benchmarks. `FTL(s)' refers to the First Token Latency (seconds). `Token$^V$' means the average number of visual tokens of a single page/frame.}
    \label{tab:multi-image}
    \footnotesize
    %\small
    \centering
    \begin{tabular}{lc|cc|cc|cc}
    \toprule
    \multirow{2}*{\textbf{Model}} & \multirow{2}*{\textbf{Token$^V$}} & \multicolumn{2}{c|}{\textbf{MP-DocVQA}} & \multicolumn{2}{c|}{\textbf{DUDE}} & \multicolumn{2}{c}{\textbf{NewsVideoQA}} \\ 
    ~ & ~ &  FTL(s)$\downarrow$ & ANLS$\uparrow$ & FTL(s)$\downarrow$  & ANLS$\uparrow$ & FTL(s)$\downarrow$   & ANLS$\uparrow$ \\
    \toprule
    LongVA-7B & $\sim$2,029 & 2.13  & 60.80 & 2.26 & 38.37  & 4.29  &  50.61 \\
    Idefics3-8B & $\sim$838 & 2.26  & 67.15 & 2.29  & 38.65 & 6.39  & 60.16 \\
    LLaVA-next-interleave-7B & 729 & 1.56  & 44.87 & 1.47  & 28.03 & 4.35  & 56.66 \\
    \modelname-8B & 324 & \textbf{0.95} & \textbf{69.42} & \textbf{0.94} & \textbf{46.77} & \textbf{1.17} & \textbf{64.09} \\
    \midrule
    \end{tabular}
\end{table*}

\begin{table*}
    \caption{Ablation study about the architecture of the compressor on single-image document benchmarks. `Img$^{base}$' refers to the basic resolution of the global image and each sub-image.}
    \label{tab:arch_ablation}
    \footnotesize
    % \small
    \centering
    \resizebox{\linewidth}{!}{
    \begin{tabular}{c|cc|ccccc|ccc|c}
    \toprule
    ~ & \multirow{2}*{\textbf{Img$^{base}$}} &  \multirow{2}*{\textbf{Crop}} & \multicolumn{5}{c|}{\textbf{Compressor}} & \multirow{2}*{\textbf{DocVQA}}  & \multirow{2}*{\textbf{WTQ}} & \multirow{2}*{\textbf{ChartQA}} \\
     ~ & ~ & ~ & Name & Compressing & Layer & Position & Token$^V$  & ~ & ~ & ~  \\
     \toprule
     r1 & 448 & 9 & Resampler & learnable query & - & after H-Reducer & 256 & 69.0  & 29.4  & 66.6 \\
     r2 & 448 & 9 & CAbstractor & Adaptive Mean & - & after H-Reducer & 256 & 73.0 & 32.6 & 67.6\\
     r3 & 448 & 9 & DocCompressor & Group Att & 2 & after H-Reducer & 256 &  \cellcolor{Gray}76.1 & \cellcolor{Gray}35.1  & \cellcolor{Gray}69.2 \\
     \midrule
     r4 & 448 & 9 & DocCompressor & Group Att & 2 & after ViT & 256 & 75.7 & 33.3  & 68.7 \\
     r5 & 448 & 9 & DocCompressor & Complete Att & 2 & after H-Reducer & 256 & 74.4 & 33.7 & 68.2 \\
     r6 & 448 & 9 & DocCompressor & Group Mean & - & after H-Reducer & 256 & 74.6 & 31.9  & 68.2 \\
     \midrule
     r7 & 448 & 9 & DocCompressor & Group Att & 1 & after H-Reducer & 256 & 76.4 & 34.2 & 69.2 \\
     r8 & 448 & 9 & DocCompressor & Group Att & 4 & after H-Reducer & 256 & 75.9 & 35.8  & 70.1 \\
     \midrule
    r9 & 448 & 12 & DocCompressor & Group Att & 2 & after H-Reducer & 256 & 76.8 & 35.6  & 69.5\\
    r10 & 504 & 12 & DocCompressor & Group Att & 2 & after H-Reducer & 324 & \cellcolor{Gray}78.7 & \cellcolor{Gray}36.7 & \cellcolor{Gray}69.4\\
     
    \midrule
    \end{tabular}
    }
\end{table*}

\subsection{Ablation Study}
We perform sufficient ablation studies to show the effectiveness of the architecture of \compresssorname~ and the three-stage training strategy of \modelname.
\subsubsection{Compressor Architecture}
To validate the effectiveness of our \compresssorname, we compare different compressing architectures with an identical training pipeline of Single-image Pretraing and Single-image Document Understanding Finetuning, keeping both training data and training setting consistent. 

As shown in \cref{tab:arch_ablation}, compared with CAbstractor~\citep{honeybee}, Resampler~\citep{qwenvl} achieves worse document understanding performance (r2 vs r1). This shows that due to no prior knowledge, such as spatial relationship, is leveraged as compressing guidance, utilizing queries learned from scratch to compress rich visually-situated text information is more challenging than simple adaptive mean pooling. Our \compresssorname~outperforms CAbstractor (r3 vs r2), validating that leveraging global visual features as layout-aware guidance can better distinguish the information density of each fine-grained visual feature and therefore maintain more visually-situated text information. 

Instead of placing the compressor after the vision-to-text module H-Reducer, we also try inserting it between the vision encoder and the vision-to-text module. Such a setting results in performance decreases across three datasets (r4 vs r3), validating our hypothesis that compressing features after the vision-to-text module is like summarizing textual features and can maintain more textual semantics while compressing visual features after the visual encoder loses more visually situated text information. Besides, without aligning each query token in the global feature map with $R \times C$ fine-grained visual tokens from the re-organized feature map to perform attention within a group as \cref{equ:group_att}, we try utilizing each query token to attend all visual tokens of sub-images. Such complete attention not only brings higher computational complexity but also causes performance decreases (r5 vs r3), showing that the positional correspondence between the global visual map and the re-organized fine-grained visual map is a reliable prior knowledge for compressing visual features efficiently. Furthermore, directly performing mean pooling on each group of $R \times C$ fine-grained visual features underperforms utilizing global visual features as the query to perform cross-attention (r6 vs r3). This also proves the importance of reliable guidance during compressing.

Compared with 2 layers of cross-attention, decreasing cross-attention layers bring a slight performance increase on DocVQA~\citep{docvqa} but more performance decrease on WikiTablesQA (WTQ)~\citep{wikitableqa} (r7 vs r3). Further increasing to 4 layers doesn't significantly improve performance (r8 vs r3). This shows that compressing high-resolution visual features doesn't require a deep neural network. Finally, increasing the maximum number of crops and the base resolution of the global image or each sub-image are two main strategies to increase the supported input resolution. Our experiments show that increasing the cropping number (r9 vs r3) or basic resolution (r10 vs r9) benefits the document understanding performance. Increasing basic resolution brings more improvement because of more visual tokens after compressing.

\begin{table*}
    \caption{Ablation study about the training stages of \modelname. `Single' and `Multi' refer to training samples utilizing single or multiple images as input. `Page Num' and `Evidence Page' refer to the number of input page images and the page ordinal number with the ground-truth answer.}
    \label{tab:train_ablation}
    \footnotesize
    % \small
    \centering
    \resizebox{\linewidth}{!}{
    \begin{tabular}{c|cc|cc|c|ccc|ccc|c}
    \toprule
     ~ & \multicolumn{2}{c|}{\textbf{Pretraining}} & \multicolumn{2}{c|}{\textbf{SFT}} & \multirow{3}*{\textbf{DocVQA}} & \multicolumn{7}{c}{\textbf{MP-DocVQA}} \\
    ~ & \multirow{2}*{\textbf{Single}} & \multirow{2}*{\textbf{Multi}} & \multirow{2}*{\textbf{Single}} & \multirow{2}*{\textbf{Multi}} & ~ & \multicolumn{3}{c|}{\textbf{Page Num}} & \multicolumn{3}{c|}{\textbf{Evidence Page}} & \multirow{2}*{\textbf{Overall}}  \\
    ~ & ~ & ~ & ~ & ~ & ~ & 1 & 2-10 & \textgreater10 & 1 & 2-10 & \textgreater10 & ~  \\
    \toprule
    r1 & \checkmark & & \checkmark & & 78.7  & 81.3 & 55.0 & 5.8 & 67.7 & 45.9 & 6.2 & 54.2 \\ 
    r2 & \checkmark & &  & \checkmark & 75.2 & 78.7 & 65.2 & 34.6 & 74.3 & 54.9 & 40.9 & 63.8 \\ 
    r3 & \checkmark & \checkmark & & \checkmark & 74.2  & 78.9 & 65.7 & 37.9 & 74.2 & 56.8 & 43.4 & 64.7  \\ 
     r4 & \checkmark & \checkmark & \checkmark  & \checkmark & \textbf{80.7} & \textbf{83.3} & \textbf{70.2} & \textbf{42.5} & \textbf{78.6} & \textbf{60.9} & \textbf{53.6} & \textbf{69.4}  \\ 
    \midrule
    \end{tabular}
    }
\end{table*}

\subsubsection{Training Strategy}
\modelname~is trained with three stages: Single-image Pretraining, Multi-image Continue-pretraining, and Multi-task Finetuning. \cref{tab:train_ablation} shows the influence of each stage for OCR-free single-page and multi-page document understanding. With the Single-image Pretraining and Single-image finetuning (r1), the model achieves promising performance on single-page benchmark DocVQA and documents from MP-DocVQA with only 1 page. Although only trained with 1 image as the input, the model can also achieve around 50\% accuracy when fed into 2-10 page images. However, the model struggles to understand documents with more than 10 pages, which greatly exceeds the number of input images during training and brings great difficulty in correlating images and finding answers. Performing Multi-image Fintuing could greatly improve the model's ability to understand multiple images (r2 vs r1). Furthermore, adding the Multi-image Continue-pretraining could also improve the question-answering performance on downstream datasets, especially for documents with more than 10 pages (r3 vs r2). This demonstrates that parsing texts of the specified page or judging which pages contain specified texts among multi-page documents is a basic ability for multi-page document understanding. Finally, by ensembling both single-image and multi-image instruction tuning sets to perform the Multi-task Finetuning (r4), \modelname~achieves the best performance on both single-page and multi-page document benchmarks, showing the cross-improvement between single-image and multi-image comprehension.

\subsection{Qualitative Results}

\begin{figure}[tp]
\begin{center}
    \includegraphics[width=1.0\linewidth]{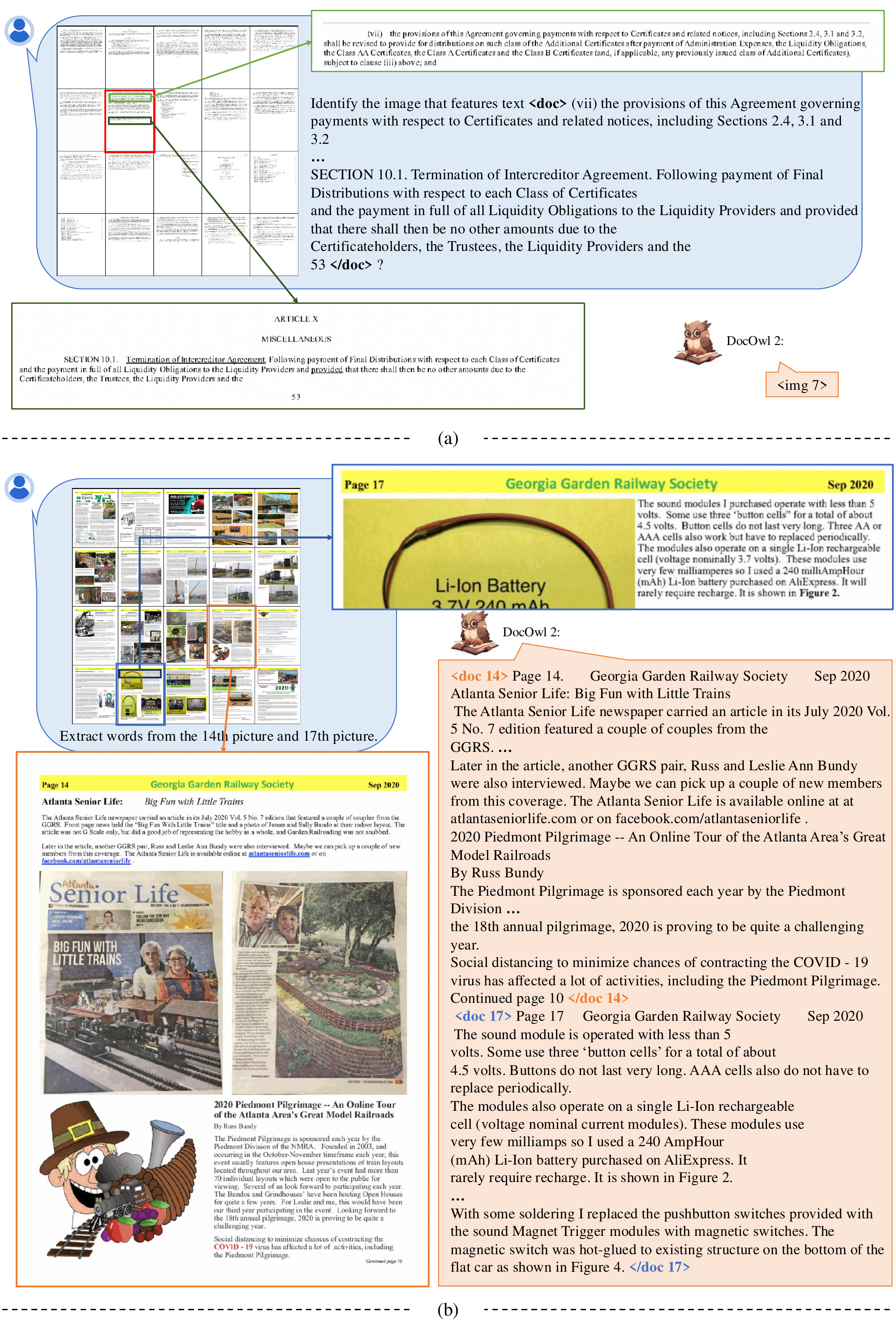}
\end{center}
\caption{Qualitative results of the Multi-page Text Lookup (a) and Multi-page Text Parsing (b) given by \modelname~after the Multi-image Continue Pretraining.}
\label{fig:mp_text_lookup_read}
\end{figure}

\begin{figure}[tp]
\begin{center}
    \includegraphics[width=1.0\linewidth]{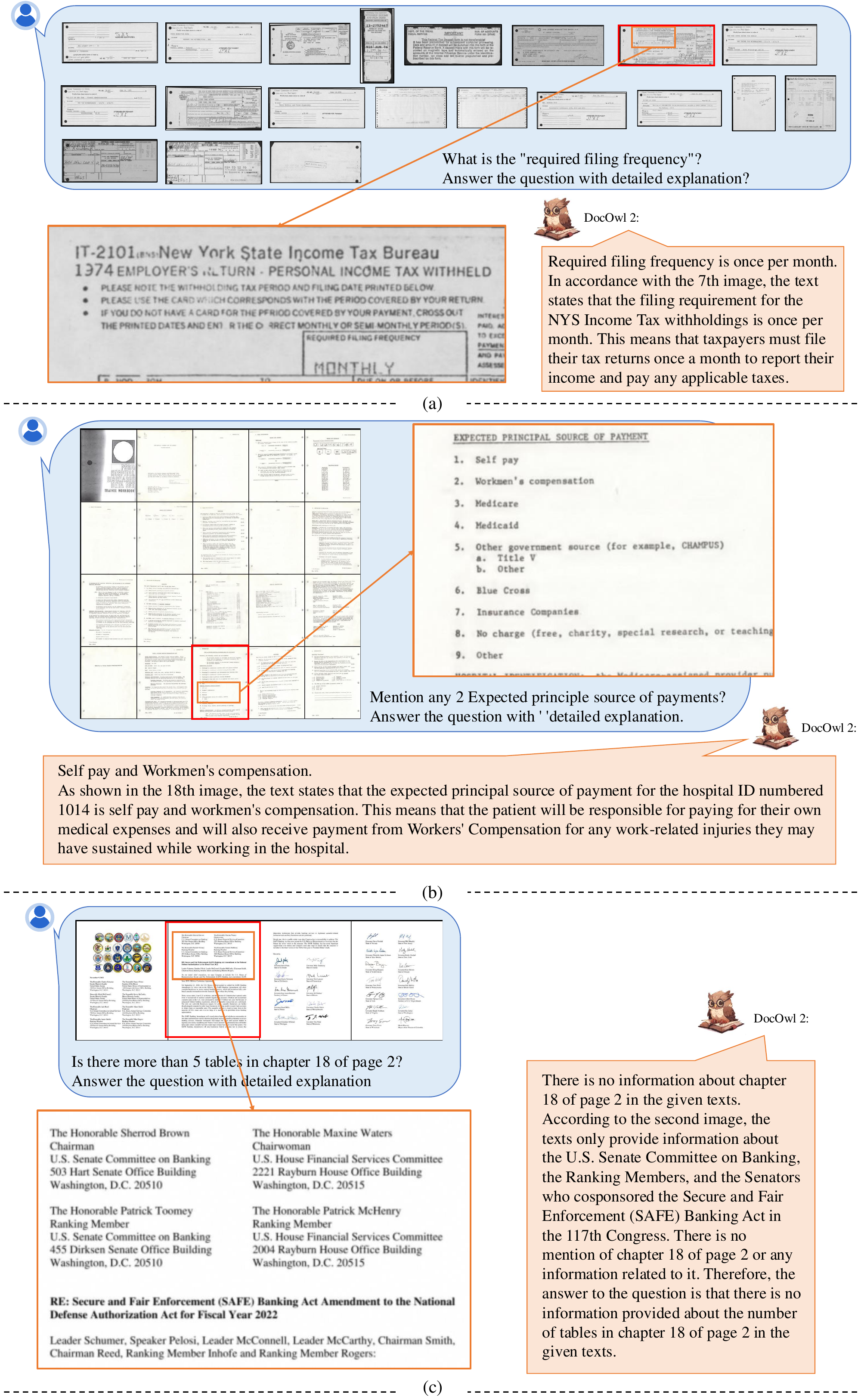}
\end{center}
\vspace{-8pt}
\caption{Qualitative results of the Multi-page Question Answering with detailed explanation.}
\label{fig:mp_qa}
\end{figure}

\begin{figure}[tp]
\begin{center}
    \includegraphics[width=1.0\linewidth]{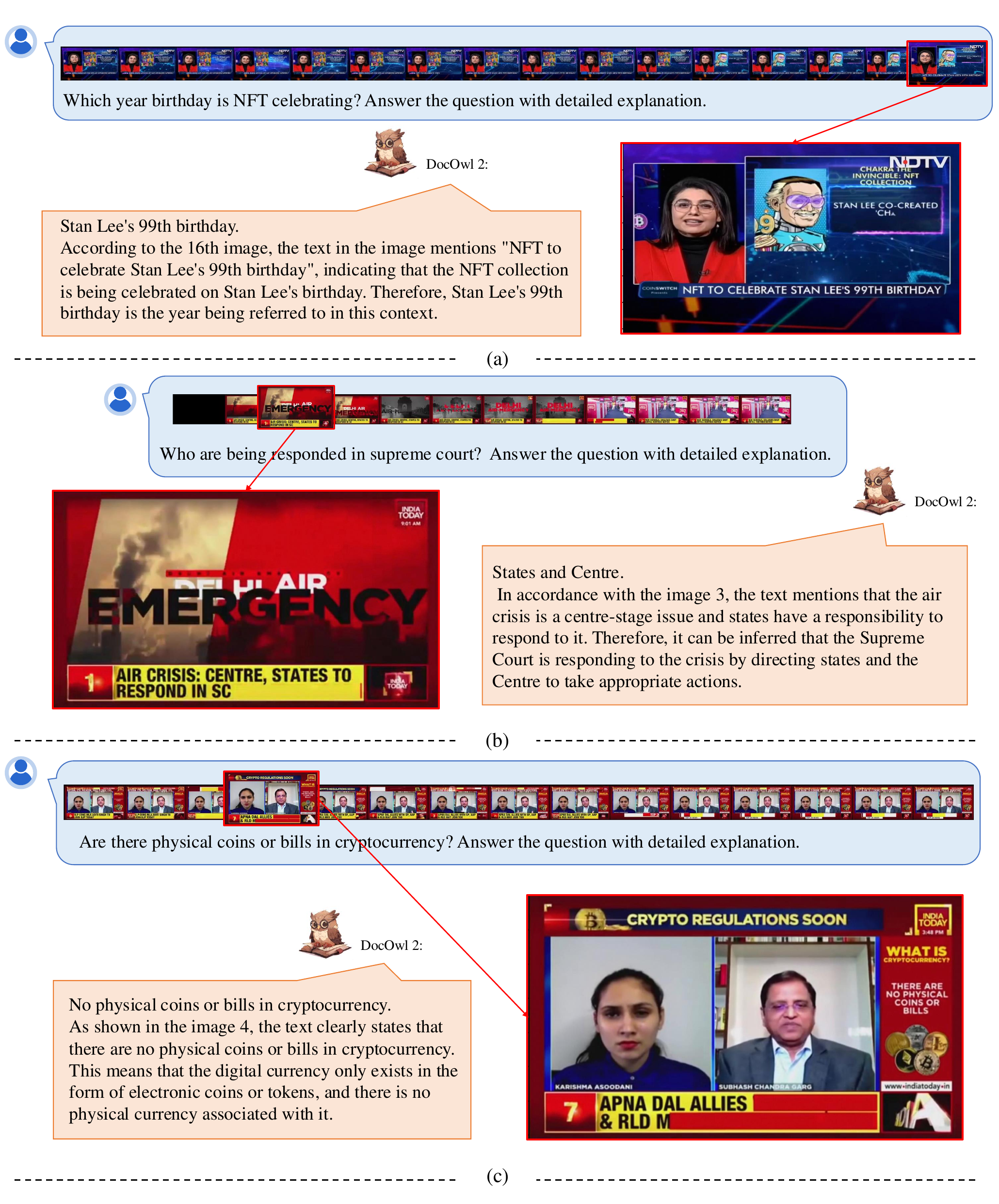}
\end{center}
\caption{Qualitative results of the Text-rich Video Understanding.}
\label{fig:video_qa}
\end{figure}

As shown in \cref{fig:mp_text_lookup_read}, after the Multi-image Continue Pretraining stage, \modelname~is able to locate the corresponding image of the given texts accurately. Besides, although representing each high-resolution image with just 324 tokens, \modelname~is still capable of parsing detailed texts of specified two images, validating the promising OCR-free multi-page document understanding performance of \modelname~. It also demonstrates our proposal that 324 tokens are enough to encode detailed text information in common A4-sized document pages and the effectiveness of our \compresssorname.

After the Multi-task Finetuning, given multiple images and a question, \modelname~can give a simple answer first and then provide a detailed explanation with the evidence, as shown in \cref{fig:mp_qa}. \modelname~can comprehend not only page images rendered from PDF files (\cref{fig:mp_qa}(c)) but also scan images of a document (\cref{fig:mp_qa}(a-b)). When a question is unanswerable, \modelname~can also tell and give corresponding reasons (\cref{fig:mp_qa}(c)). 

Besides multi-page documents, \modelname~is also capable of understanding text-rich videos. As shown in \cref{fig:video_qa}, among similar frames within a video, \modelname~can distinguish fine-grained textual differences, locate relevant frames, and give accurate answers.

%% file: Conclusion.tex
\section{Conclusion}
In this work, we propose mPLUG-\modelname, a Multimodal Large Language Model with the ability of efficient OCR-free Multi-page Document Understanding. The novel architecture~\compresssorname~in \modelname~compresses each high-resolution document image into 324 tokens through cross-attention with the global visual feature as guidance, and re-organized features of cropped images as keys and values. On single-image document understanding benchmarks, with fewer visual tokens, \modelname~outperforms existing compressing methods and achieves comparable performance with SOTA MLLMs with similar training data. Besides, \modelname~achieves OCR-free state-of-the-art performance on two multi-page document understanding benchmarks and 1 text-rich video understanding benchmark. Our experiments validate that thousands of visual tokens for 1 common A4-sized document page may be so redundant that too many computational resources are wasted. We hope \modelname~could bring more researchers' attention to the balance of efficient representation of high-resolution images and OCR-free Document Understanding performance.